\documentclass[runningheads]{llncs}
\usepackage[T1]{fontenc}
\usepackage{graphicx}
\usepackage{booktabs}
\usepackage{array}
\usepackage{tabularx}
\usepackage{booktabs}
\usepackage{amsmath} 
\usepackage{multirow}
\usepackage{threeparttable}
\usepackage{orcidlink}

\newcolumntype{L}[1]{>{\raggedright\arraybackslash}m{#1}} 
\newcolumntype{C}[1]{>{\centering\arraybackslash}m{#1}}

\begin{document}

\title{Human--AI Collaboration Reconfigures Group Regulation from Socially Shared to Hybrid Co-Regulation}
\titlerunning{Human--AI Collaboration Reconfigures Group Regulation}

\author{
Yujing Zhang\inst{1}\orcidlink{0009-0009-3324-2501} \and
Xianghui Meng\inst{1}\orcidlink{0009-0000-9421-8243} \and
Shihui Feng\inst{1}\orcidlink{0000-0002-5572-276X} \and
Jionghao Lin\inst{1\thanks{Corresponding author.},2,3}\orcidlink{0000-0003-3320-3907}
}

\authorrunning{Y. Zhang et al.}

\institute{
The University of Hong Kong, Hong Kong SAR, China \\
\email{{\{u3011372, u3014272\}@connect.hku.hk}}\\
\email{{\{shihuife, jionghao\}@hku.hk}}
\and
Carnegie Mellon University, Pittsburgh, PA, USA
\and
Monash University, Clayton, VIC, Australia
}

\maketitle

\begin{abstract}

Generative AI (GenAI) is increasingly used in collaborative learning, yet its effects on how groups regulate collaboration remain unclear. Effective collaboration depends not only on what groups discuss, but on how they jointly manage goals, participation, strategy use, monitoring, and repair through co-regulation and socially shared regulation. We compared collaborative regulation between Human--AI and Human--Human groups in a parallel-group randomised experiment with 71 university students completing the same collaborative tasks with GenAI either available or unavailable. Focusing on human discourse, we used statistical analyses to examine differences in the distribution of collaborative regulation across regulatory modes, regulatory processes, and participatory focuses. Results showed that GenAI availability shifted regulation away from predominantly socially shared forms towards more hybrid co-regulatory forms, with selective increases in directive, obstacle-oriented, and affective regulatory processes. Participatory-focus distributions, however, were broadly similar across conditions. These findings suggest that GenAI reshapes the distribution of regulatory responsibility in collaboration and offer implications for the human-centred design of AI-supported collaborative learning.

\end{abstract}

\section{Introduction}

Generative artificial intelligence (GenAI) in computer-supported collaborative learning (CSCL) is increasingly conceptualised not only as a tool, but also as an interactive contributor that can generate ideas, challenge assumptions, and extend collective reasoning during group work \cite{yan2024genai}. However, effective collaboration depends not simply on participation, but on how joint activity is organised through shared planning, monitoring, evaluation, and adaptation over time \cite{rogat2011ssrl}. In collaborative contexts, regulation extends beyond individual self-regulated learning to include co-regulation (CoRL) and socially shared regulation (SSRL), both of which are central to effective collaboration because they rely on coordination and shared responsibility among group members \cite{hadwin2018regulation}.

Despite their importance, empirical evidence on how GenAI relates to CoRL and SSRL remains limited \cite{kim-2025}. This gap is increasingly important as GenAI is integrated into collaborative learning settings without closely examining group regulation. Recent studies suggest that GenAI can reshape participation and interaction in collaborative problem solving \cite{feng2025group}, raising concerns about how regulatory responsibility is redistributed when GenAI enters group work. Addressing this gap is therefore important for advancing Human--AI collaboration theory and informing human-centred AI design in education.

To examine whether GenAI availability reconfigures collaborative regulation, this study designed and deployed a conversational GenAI agent explicitly oriented towards supporting collaborative regulation. Rather than providing direct solutions, the agent intervened through dialogue to shape how groups coordinated and regulated their joint activity. By comparing Human--AI and Human--Human groups working on the same tasks, we examined differences in the distribution of collaborative regulation across regulatory modes, regulatory processes, and participatory focuses. Specifically, we asked: \emph{What differences are observed in the distribution of collaborative regulation between Human--AI and Human--Human groups?}

\section{Related Work}
\subsection{Co-Regulation and Socially Shared Regulation in CSCL}
When collaborative tasks become complex and open-ended, groups must regulate their activity through co-regulated learning (CoRL), and socially shared regulation of learning (SSRL) \cite{hadwin2018regulation}. CoRL and SSRL are interactional and enacted through dialogue. CoRL describes peer-supported regulation (e.g., prompts, guidance, feedback) that helps individuals manage their learning, while SSRL involves the group’s deliberate, collective regulation of goals, strategies, monitoring, and reflection. CoRL often serves as a bridge that enables SSRL to emerge and stabilise \cite{hadwin2018regulation}. As such, SSRL is widely treated as a hallmark of effective collaboration, capturing moments when groups jointly frame the task, negotiate approaches, track progress, and adapt to challenges \cite{hadwin2018regulation,jarvenoja2015understanding}.

In CSCL, however, these forms of regulation rarely arise reliably without support. Learners bring differing goals, knowledge, and affective states, so simply placing them together does not ensure the development of CoRL or SSRL \cite{lyons-2020}. Productive collaboration requires ongoing alignment, including shared task understanding, role coordination, progress monitoring, and expectation management, and these regulatory demands shift as interaction unfolds \cite{jarvela-2016}. This makes scaffolding CoRL and SSRL a central design problem in CSCL and a key target for interventions intended to improve collaborative process and outcomes \cite{edwards2025human}.

\subsection{GenAI Agent in CSCL} 
Recent advances in GenAI have opened new possibilities for CSCL interventions: beyond information delivery, GenAI systems can generate ideas, pose questions, and scaffold reasoning, suggesting potential roles in shaping collaborative learning processes \cite{bansal2024transforming}. Yet, despite the central role of co-regulation and socially shared regulation in effective CSCL, few studies have examined whether GenAI can meaningfully shape collaborative regulation in group learning. A notable early study introduced a GenAI agent to prompt metacognitive reflection in support of SSRL, but found limited evidence of improved regulation when assessed through linguistic alignment \cite{edwards2025human}. Crucially, we argue that this reflects an analytic limitation rather than a failure of AI support: treating regulation as a static shift in language use is insufficient to capture its distributed and emergent nature. Building on this insight, we treat CoRL and SSRL as interactional phenomena that can be observed through the distribution of regulatory modes, regulatory processes, and participatory discourse moves in group interaction.
\vspace{-2mm}
\section{Method}
\vspace{-1mm}
\subsection{Data Collection}
\begin{figure}[h]
    \vspace{-3mm}
    \centering
    \includegraphics[width=0.8\linewidth]{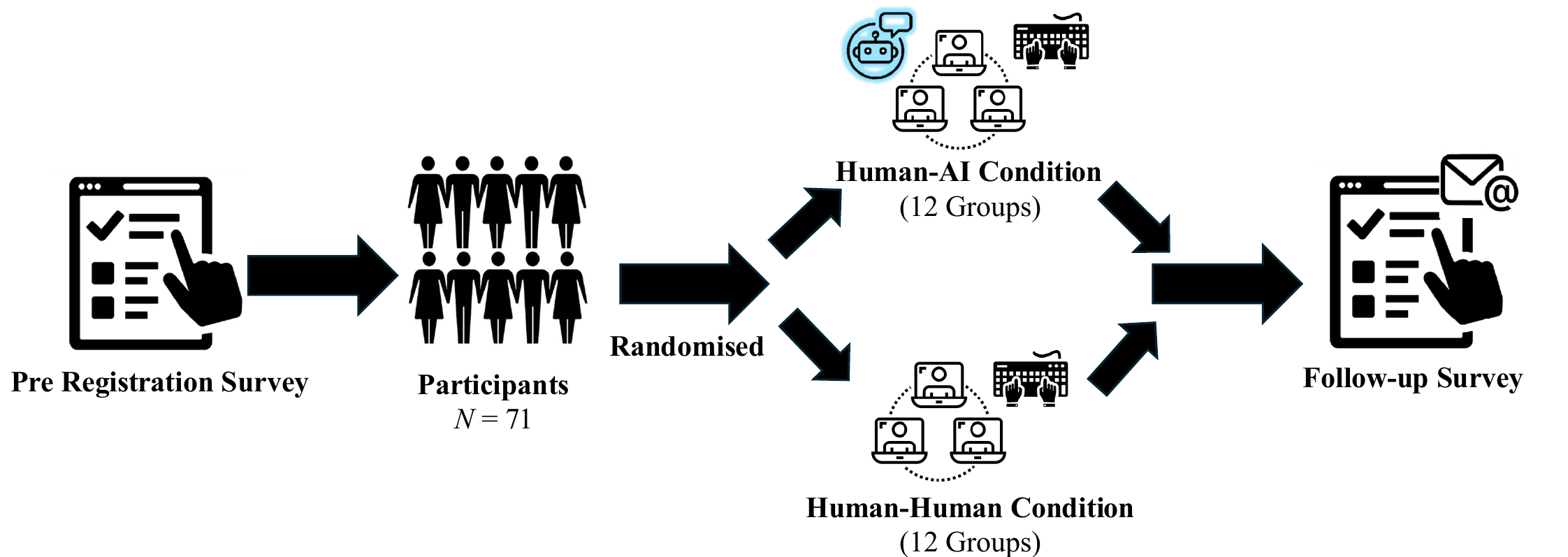} 
    \caption{Study Procedure.}
    \label{fig:procedure}
    \vspace{-9mm}
\end{figure}

Figure~\ref{fig:procedure} summarises the data collection procedure. We used a between-groups experimental design: participants were randomly assigned to a GenAI-available or GenAI-unavailable condition and then formed into small groups within condition to complete the same collaborative tasks. The sample comprised 71 university students (aged 18–35; $M=23.73$, $SD=3.65$) from English-speaking countries and was predominantly Asian ($n=66$; mixed ethnicity $n=2$; White $n=3$). Participants were allocated to 24 groups (23 triads and one dyad). The study received ethical approval from the Human Research Ethics Committee of The University of Hong Kong (Ref.\ No.:~EA250843).

\subsection{Group Task Design}
The group tasks were grounded in the collective intelligence paradigm, which posits that group performance reflects a stable collective intelligence factor rather than task-specific skills alone \cite{Woolley2010}. To capture different coordination and perspective-taking demands, we used two complementary tasks: a moral-reasoning case on academic misconduct that required value-based discussion and justified consensus among multiple-choice options, and a shopping-planning task that required integrating constraints and producing a concrete route plan.

\subsection{GenAI Agent Design}
\begin{figure}[h]
    \centering
    \includegraphics[width=\linewidth]{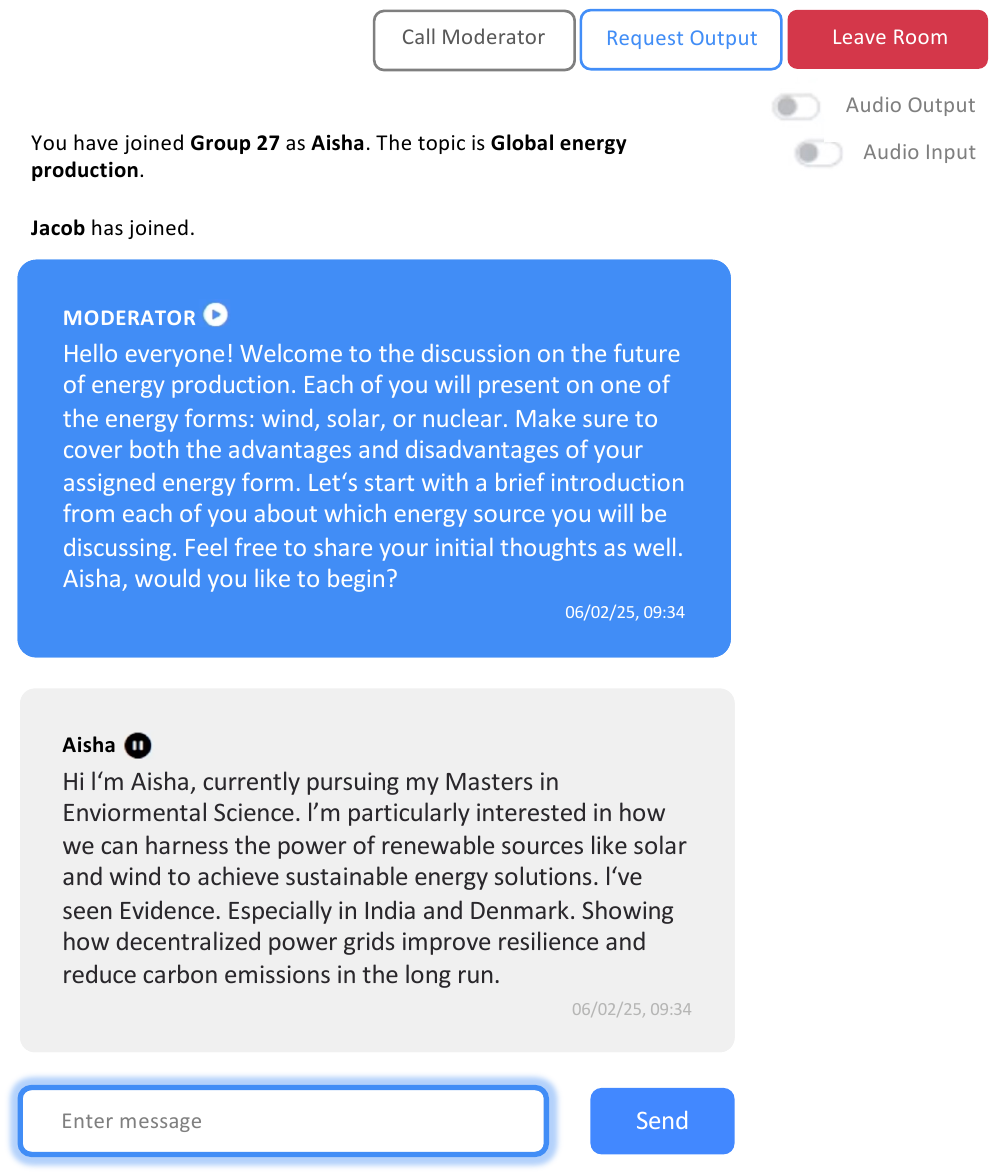} 
    \caption{GenAI agent interface embedded in the platform.}
    \label{fig:chatbot_interface}
    \vspace{-9mm}
\end{figure}

The GenAI agent (\texttt{OpenAI GPT-4o mini}) was embedded in an online CSCL platform to support real-time, text-based group collaboration (see Fig.\ref{fig:chatbot_interface}). Table~\ref{tab:genai_intervention} summarises the conversational prompts used by the GenAI agent to support collaborative regulation during group work. Rather than providing content or solutions, these prompts were designed to elicit explanation, encourage perspective-taking, and guide groups through progression of their joint activity.

\begin{table}[h]
\centering
\caption{GenAI Prompts Used to Support Collaborative Regulation}
\label{tab:genai_intervention}

\begingroup
\renewcommand{\arraystretch}{1.1}
\setlength{\tabcolsep}{6pt}
\resizebox{0.88\textwidth}{!}{
\begin{tabularx}{\linewidth}{@{}
>{\raggedright\arraybackslash}p{4.2cm}
>{\raggedright\arraybackslash}X
@{}}
\toprule
\textbf{Intervention Behaviour} & \textbf{Example Prompt} \\
\midrule

Further Reasoning &
\textit{“Could you explain your answer in more detail?”} \\
\midrule

Invite Examples &
\textit{“Can you give an example to support your point?”} \\
\midrule

Challenge Viewpoints &
\textit{“Do you think this always holds, or might there be exceptions?”} \\
\midrule

Turn-Taking Management &
\textit{“Thank you Jack! Anyone else want to share their view?”} \\
\midrule

Summarising &
\textit{“Let me summarise the group discussion.”} \\
\midrule

Task Transition &
\textit{“How would you like to move forward from here?”} \\
\midrule

Invite Building on Ideas &
\textit{“How does this connect with what was mentioned earlier?”} \\
\bottomrule
\end{tabularx}
}
\endgroup
\end{table}

\subsection{Measure}

\subsubsection {Group Regulatory Dynamics}
Group regulatory dynamics were analysed using a two-level coding framework that distinguishes extended regulatory activity from moment-to-moment discourse actions.

At the episode level, a regulatory episode was defined as a continuous segment focused on a single regulatory issue, typically spanning multiple turns. Episodes began when a regulatory concern was introduced and ended when it was resolved or the group shifted focus. Each episode was coded for regulatory mode: CoRL, SSRL, or hybrid CoRL\&SSRL (i.e., individual guidance and shared regulation co-occurring within the same episode). Episodes were then coded for specific regulatory processes (adapted from Nguyen et al.\cite{nguyen-2022}; see Table~\ref{tab:episode_regulation_process}).

At the utterance level, each conversational turn was coded for participatory focus (adapted from Dang et al.\cite{dang-2024}; see Table~\ref{tab:utterance_regulatory_focus}). This captures the speaker’s immediate regulatory action(e.g., planning next steps, reasoning about options, reporting outcomes) \cite{FengWei2025CoMPAS}. 

\begingroup
\renewcommand{\arraystretch}{1.1}
\setlength{\tabcolsep}{3pt}
\setlength{\abovecaptionskip}{1pt}
\setlength{\belowcaptionskip}{0pt}

\begin{table}[h]
\centering
\caption{Episode-Level Regulatory Processes}
\label{tab:episode_regulation_process}

\begin{threeparttable}
\resizebox{\textwidth}{!}{%
\begin{tabular}{L{1.2cm} L{3.9cm} L{5.2cm} L{4.3cm}}
\toprule
\textbf{Code} & \textbf{Subcode} & \textbf{Operational definition} & \textbf{Example} \\
\midrule

\multirow{3}{*}{CoRL}
& Obstacle Detection
& Identifying an obstacle in another member’s learning or task progress.
& ``\textit{You seem stuck on this step.}'' \\
\cline{2-4}

& Strategic Action Direction
& Directing another member to adopt or change a strategy.
& ``\textit{Jack, you can use the formula instead.}'' \\
\cline{2-4}

& Affective Support
& Regulating another member’s emotional or motivational state.
& ``\textit{Don’t worry, Lily!}'' \\
\midrule

\multirow{4}{*}{SSRL}
& Shared Obstacle Negotiation
& Jointly identifying and discussing group-level obstacles.
& \textbf{A}: ``\textit{We’re off track.}'' \newline
  \textbf{B}: ``\textit{Yeah, we haven’t answered the question.}'' \\
\cline{2-4}

& Shared Strategic Negotiation
& Jointly proposing and negotiating alternative strategies.
& \textbf{A}: ``\textit{Can we compare both cases?}'' \newline
  \textbf{B}: ``\textit{Yeah, let’s start with the first one.}'' \\
\cline{2-4}

& Shared Action Change
& Jointly agreeing on a change of action and coordinating execution.
& \textbf{A}: ``\textit{Let’s change our method.}'' \newline
  \textbf{B}: ``\textit{Okay, let’s do that.}'' \\
\cline{2-4}

& Shared Affective Regulation
& Jointly regulating the group’s emotional or motivational state.
& \textbf{A}: ``\textit{This is frustrating.}'' \newline
  \textbf{B}: ``\textit{Yeah, but we can calm down and keep going.}'' \\
\bottomrule
\end{tabular}
}

\begin{tablenotes}[flushleft]
\footnotesize
\item[] \textit{Note.} \textbf{A} and \textbf{B} denote different human group members. 
\end{tablenotes}
\vspace{-4mm}
\end{threeparttable}
\end{table}

\endgroup

\begingroup
\renewcommand{\arraystretch}{1.1}
\setlength{\tabcolsep}{3pt}
\setlength{\abovecaptionskip}{1pt}
\setlength{\belowcaptionskip}{0pt}

\begin{table}[h]
\centering
\caption{Utterance-Level Participatory Focus}
\label{tab:utterance_regulatory_focus}
\resizebox{\textwidth}{!}{%
\begin{tabular}{L{2.4cm} L{2.4cm} L{4cm} L{4.3cm}}
\toprule
\textbf{Code} & \textbf{Subcode} & \textbf{Operational definition} & \textbf{Example} \\
\midrule

\multirow{3}{*}{Cognitive}
& Cog--Explain
& Explains task concepts or rules to build shared understanding.
& ``\textit{Frozen food must be returned within 40 minutes.}'' \\
\cline{2-4}

& Cog--Reason
& Uses reasoning to justify choices or compare options.
& ``\textit{If we exceed \pounds 50, we lose points.}'' \\
\cline{2-4}

& Cog--Evaluate
& Judges the quality or fairness of an option.
& ``\textit{Choice A is too harsh.}'' \\
\midrule

\multirow{4}{*}{Metacognitive}
& Meta--Orient
& Frames what the group should decide or focus on.
& ``\textit{Let’s start with Question 1.}'' \\
\cline{2-4}

& Meta--Plan
& Proposes strategies, sequencing, or next steps.
& ``\textit{Everyone give your answer, then we compare.}'' \\
\cline{2-4}

& Meta--Monitor
& Checks progress, time, understanding, or agreement.
& ``\textit{How much time do we have left?}'' \\
\cline{2-4}

& Meta--Reflect
& Evaluates past strategies or collaboration quality.
& ``\textit{This approach isn’t working well.}'' \\
\midrule

\multirow{2}{*}{Task execution}
& TE--Act
& Indicates execution of a concrete task action.
& ``\textit{I’ll calculate this part.}'' \\
\cline{2-4}

& TE--Report
& Reports intermediate or final task results.
& ``\textit{The answer is 175.}'' \\
\midrule

\multirow{2}{*}{Socio-emotional}
& SE--Express
& Expresses an emotional reaction.
& ``\textit{This is funny lol.}'' \\
\cline{2-4}

& SE--Regulate
& Regulates group emotion or motivation.
& ``\textit{Good job guys!}'' \\
\bottomrule
\end{tabular}
}
\vspace{-3mm}
\end{table}
\endgroup

Two trained coders independently annotated the data. Inter-rater agreement was near perfect for utterance-level participatory focus ($\kappa = .85$–$.98$) and episode-level regulatory processes ($\kappa = .86$–$.95$), with high span-based episode segmentation accuracy (macro F1 = .94, boundary F1 = .97). This span-level evaluation was adopted because regulatory episodes vary in length, and reliable coding requires both consistent category assignment and accurate identification of episode boundaries.

\section{Data Analysis}

All analyses were conducted at the group level; in the Human--AI condition, GenAI-generated turns were excluded so that findings reflected how GenAI availability reshaped human regulatory organisation rather than effects driven by GenAI utterance volume or content.

We used mixed-design ANOVAs because category (i.e., regulatory mode, regulatory process, or participatory focus) was a within-group factor measured repeatedly, whereas condition (i.e., Human--AI vs Human--Human) was a between-group factor. Post hoc Holm-adjusted pairwise contrasts were then used to compare conditions within each category while controlling the family-wise error rate across multiple tests.

\section{Results}

\subsection{Human--AI Groups Shifted Towards Hybrid Co-regulatory Organisation}

\subsubsection{Episode-level regulation mode.}
A mixed-design ANOVA on regulatory-mode proportions revealed a significant main effect of mode, $F(1.20, 26.31)=89.28, p<.001$, indicating that regulatory responsibility was unevenly distributed across modes. There was no main effect of condition, $F(1,22)\approx0.00, p>.999$, but a significant \textit{Condition × Mode} interaction, $F(1.20,26.31)=222.46, p<.001$, indicating that regulatory organisation differed between conditions. Holm-adjusted contrasts (see Table~\ref{tab:episode_regulation_distribution}) showed that Human–Human groups exhibited a higher proportion of SSRL episodes, whereas Human–AI groups showed higher proportions of hybrid CoRL\&SSRL and CoRL-only episodes. Overall, GenAI availability was associated with a shift from predominantly socially shared regulation towards more hybrid and co-regulatory organisation.
\vspace{-3mm}









\begin{table}[ht]
\centering
\caption{Distribution of regulation modes by condition}
\label{tab:episode_regulation_distribution}

\begingroup
\renewcommand{\arraystretch}{0.8} 
\setlength{\tabcolsep}{2pt}
\setlength{\abovecaptionskip}{2pt}
\setlength{\belowcaptionskip}{0pt}
\setlength{\aboverulesep}{0.2ex}
\setlength{\belowrulesep}{0.2ex}

\begin{threeparttable}
\begin{tabularx}{\textwidth}{@{}
>{\raggedright\arraybackslash}m{3.0cm}
>{\centering\arraybackslash}X
>{\centering\arraybackslash}X
@{\hspace{18pt}} 
>{\hspace{3pt}\raggedleft\arraybackslash}m{2cm} 
>{\raggedright\arraybackslash}m{2.2cm} 
@{}}
\toprule
\shortstack{\textbf{Regulation mode}} &
\shortstack{\textbf{Human--AI}} &
\shortstack{\textbf{Human--Human}} &
\shortstack{\textbf{t statistic}\\\textbf{(df = 22)}} \\
\cmidrule(lr){2-3}
& Mean Prop.\ (SD) & Mean Prop.\ (SD) & \\
\midrule
CoRL         & 0.066 (0.08) & 0.000 (0.00) & $-2.87^{**}$  \\
SSRL         & 0.083 (0.11) & 0.940 (0.12) & $18.29^{***}$ \\
CoRL\&SSRL & 0.851 (0.16) & 0.060 (0.12) & $-13.54^{***}$ \\
\bottomrule
\end{tabularx}

\begin{tablenotes}[flushleft]
\footnotesize
\item[] \textit{Note.} Mean Prop.\ indicates the average proportion of a group’s total regulatory episodes accounted for by each mode.
$^{***}p < .001$; $^{**}p < .01$.
\end{tablenotes}
\end{threeparttable}
\endgroup
\vspace{-3mm}
\end{table}

\subsubsection{Episode-level regulation process.}
A mixed-design ANOVA on regulatory-process proportions revealed significant main effects of condition, $F(1,22)=98.07, p<.001$, and process category, $F(3.56,78.23)=63.88, p<.001$, as well as a significant \textit{Condition × Process} interaction, $F(3.56,78.23)=33.53, p<.001$, indicating process-specific condition differences. Holm-adjusted contrasts (Table~\ref{tab:episode_subcode_prop_distribution}) showed that Human–AI groups had higher proportions of \textit{Affective Support}, \textit{Obstacle Detection}, \textit{Shared Affective Regulation}, and \textit{Strategic Action Direction}. Overall, GenAI availability was associated with a selective reconfiguration of regulatory processes rather than a uniform shift across functions.

\begin{table}[ht]
\centering
\caption{Distribution of Episode-Level Regulatory Processes by Condition}
\label{tab:episode_subcode_prop_distribution}

\begingroup
\renewcommand{\arraystretch}{0.8} 
\setlength{\tabcolsep}{2pt}
\setlength{\abovecaptionskip}{2pt}
\setlength{\belowcaptionskip}{0pt}
\setlength{\aboverulesep}{0.2ex}
\setlength{\belowrulesep}{0.2ex}

\begin{threeparttable}
\begin{tabularx}{\textwidth}{@{}
>{\raggedright\arraybackslash}m{4.5cm}
>{\centering\arraybackslash}m{2.7cm}
>{\centering\arraybackslash}m{2.7cm}
@{\hspace{10pt}}   
>{\hspace{4pt}\raggedright\arraybackslash}m{2cm} 
>{\raggedright\arraybackslash}m{3 cm} 
@{}}
\toprule

\shortstack{\textbf{Regulatory process}} &
\shortstack{\textbf{Human--AI}} &
\shortstack{\textbf{Human--Human}} &
\shortstack{\textbf{t statistic}\\\textbf{(df = 22)}} \\
\cmidrule(lr){2-3}
& Mean Prop.\ (SD) & Mean Prop.\ (SD) & \\
\midrule
Affective Support    & 0.837 (0.20) & 0.00 (0.00) & $-14.62^{***}$  \\
Obstacle Detection & 0.314 (0.21) & 0.183 (0.21) & $-2.20^{*}$ \\
Strategic Action Direction & 0.808 (0.21) & 0.060 (0.12) & $-10.58^{***}$ \\
Shared Obstacle Negotiation & 0.284 (0.17) & 0.394 (0.19) & $1.26$ \\
Shared Strategic Negotiation & 0.829 (0.18) & 0.824 (0.17) & $-0.07$ \\
Shared Action Change & 0.817 (0.20) & 0.778 (0.22) & $-0.46$ \\
Shared Affective Regulation & 0.838 (0.22) & 0.128 (0.18) & $-9.14^{***}$ \\

\bottomrule
\end{tabularx}

\begin{tablenotes}[flushleft]
\footnotesize
\item[] \textit{Note.} Mean Prop. indicates the average proportion of a group’s total regulatory episodes accounted for by each process category. 
$^{***}p < .001$; $^{*}p < .05$.
\end{tablenotes}
\end{threeparttable}
\endgroup

\end{table}

\subsubsection{Utterance-level participatory focus.}
A mixed-design ANOVA on participatory-focus proportions showed a significant main effect of focus category, $F(3.56,78.41)=92.72, p<.001$, and a small main effect of condition, $F(1,22)=4.70, p=.041$, but no \textit{Condition × Focus} interaction, $F(3.56,78.41)=1.44, p=.233$. Holm-adjusted contrasts revealed no reliable condition differences within any focus category (all $p \ge .0646$). This pattern suggests that although overall participatory focus distributions differed slightly between conditions, GenAI availability did not selectively shift attention toward any specific focus type.

\section{Discussion and Conclusion}

This study provides empirical evidence that GenAI availability is associated with changes in the distribution of collaborative regulation in CSCL. Compared with Human--Human groups, Human--AI groups showed less predominantly socially shared regulation and more hybrid co-regulatory forms, together with higher proportions of directive, obstacle-oriented, and affective regulatory processes. Participatory-focus distributions, however, were broadly similar across conditions, suggesting that GenAI was associated more with a redistribution of regulatory responsibility than with broad changes in moment-to-moment discourse. These findings suggest that GenAI may reconfigure collaborative regulation by changing how groups coordinate support, monitoring, and shared control. For practice, they imply that educational GenAI should be designed to scaffold rather than replace shared regulation during collaboration.

This study has several limitations. The sample was relatively small and demographically skewed, many participants used English as a second language, and the study focused on text-only interaction. Future research should examine whether similar patterns emerge in more diverse and multimodal collaborative settings.

\section*{Acknowledgment}

This work was supported by the Faculty Research Fund and by the grant from the URC (Grant No. 2401102970) at The University of Hong Kong. 

\bibliographystyle{splncs04}
\bibliography{references}

\end{document}